\documentclass[letterpaper, 10 pt, conference]{ieeeconf}  

\IEEEoverridecommandlockouts                              

\usepackage{amsmath} 
\usepackage{amssymb}  
\usepackage{cite}
\usepackage{graphicx}
\usepackage{comment}
\usepackage[font=footnotesize]{caption}
\usepackage[subrefformat=parens]{subcaption}
\usepackage{multirow}
\usepackage{bbm}
\usepackage{dsfont}
\usepackage{color}
\newtheorem{dfn}{Definition}
\newtheorem{plm}{Problem}

\title{\LARGE \bf
Reinforcement Learning of Action and Query Policies with LTL Instructions under Uncertain Event Detector
}

\author{Wataru Hatanaka$^{1,2}$, Ryota Yamashina$^{1}$, and Takamitsu Matsubara$^{2}$
\thanks{$^{1}$Wataru Hatanaka and Ryota Yamashina are with Digital Strategy Division, RICOH Company,Ltd, Japan.
        {\tt\small \{wataru.hatanaka, ryohta.yamashina\}@jp.ricoh.com}}%
\thanks{$^{2}$Wataru Hatanaka and Takamitsu Matsubara are with Division of Information Science, Graduate School of Science and Technology, Nara Institute of Science and Technology (NAIST), Japan.
        {\tt\small takam-m@is.naist.jp}}%
}

\begin{document}

\maketitle
\thispagestyle{empty}
\pagestyle{empty}

\begin{abstract}
Reinforcement learning (RL) with linear temporal logic (LTL) objectives can allow robots to carry out symbolic event plans in unknown environments.
Most existing methods assume that the event detector can accurately map environmental states to symbolic events; however, uncertainty is inevitable for real-world event detectors.
Such uncertainty in an event detector generates multiple branching possibilities on LTL instructions, confusing action decisions.
Moreover, the queries to the uncertain event detector, necessary for the task's progress, may increase the uncertainty further. 
To cope with those issues, we propose an RL framework, Learning Action and Query over Belief LTL (LAQBL), to learn an agent that can consider the diversity of LTL instructions due to uncertain event detection while avoiding task failure due to the unnecessary event-detection query.
Our framework simultaneously learns 1) an embedding of belief LTL, which is multiple branching possibilities on LTL instructions using a graph neural network, 2) an action policy, and 3) a query policy which decides whether or not to query for the event detector. Simulations in a 2D grid world and image-input robotic inspection environments show that our method successfully learns actions to follow LTL instructions even with uncertain event detectors. 
\end{abstract}

\section{INTRODUCTION}
Service robots are expected to reduce human workload and improve the quality of human life in indoor and outdoor fields.
These robots must follow diverse instructions and perform stable, even in long-horizon tasks, to coexist with human society.
Learning these abilities, especially under the partial observability of the real world, is one of the challenges of co-working with humans autonomously.


Linear Temporal Logic (LTL) \cite{Pnueli1977-ug}, a formal language that captures the temporal property of the task as a symbolic event representation, is widely used to construct systems that satisfy desired specifications.
In recent years, reinforcement learning (RL) to maximize the probability of satisfying LTL instructions allows an agent to follow various instructions even when the environmental model is unknown \cite{toro2018teaching,camacho2019ltl,Bozkurt2021,leon2020systematic, kuo2020encoding, Vaezipoor2021-qm}.
However, the success of these methods is based on the unrealistic assumption that the event detector, which informs the agent of symbolic events occurring in the environment, does not fail to detect events and that the agent is always aware of the task's progress. 

\begin{figure}[t]
    \centering
    \includegraphics[width=84mm]{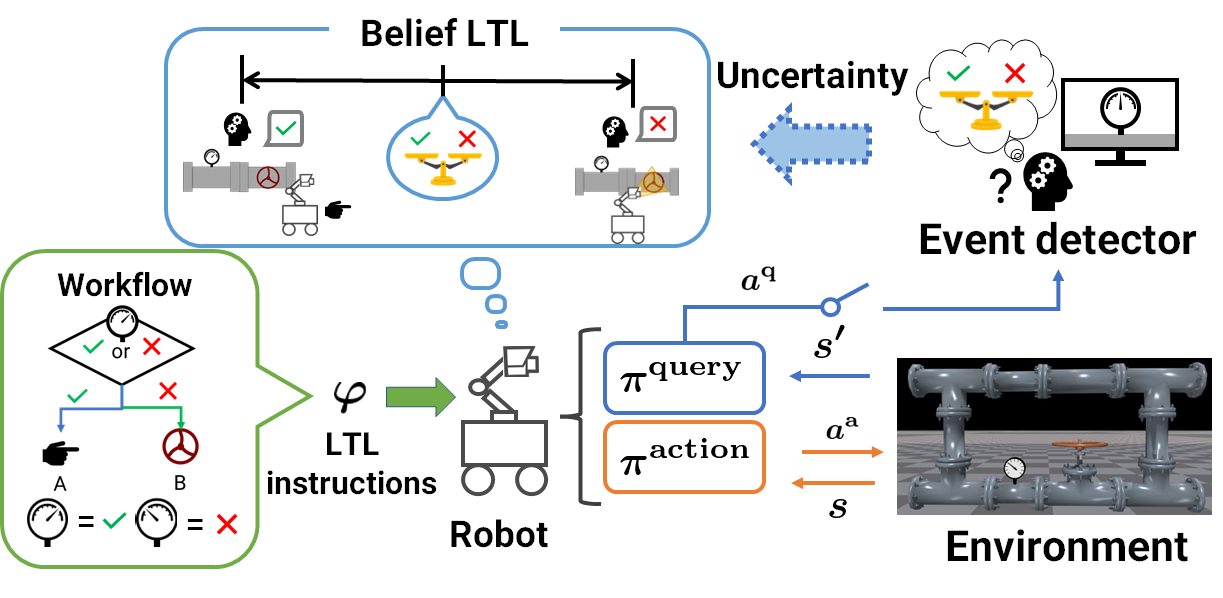}
    \vspace{-3mm}
    \caption{Overview of our proposed framework. A robot has a query policy that controls the interaction with an event detector in addition to an action policy. Our method learns both policies that act according to the belief over LTL instructions due to uncertainty in the event detector.}
    \label{fig:overview_proposed}
    \vspace{-6mm}
\end{figure}

When assuming uncertainty in event detection, we must face the risk of failure caused by giving the detection possibility to an event that has not occurred.
For example, as shown in Fig. \ref{fig:overview_proposed}, consider the situation when the inspection result contains uncertainty in a workflow that branches depending on whether the equipment is normal or abnormal.
A gradual evaluation according to the confidence or prediction score of event detection is a common approach to convey uncertainty in decision-making to the agent.
However, believing only the most likely event, employed by some methods that deal with probabilistic event detectors \cite{shah2020planning,ghasemi2020task}, may fail the task with half the probability.
To avoid this, the agent needs to act in response to different levels of uncertainty: if the uncertainty is low, the agent will believe in its judgment and act; otherwise, it will act conservatively.
Furthermore, the agent may suffer from this uncertainty each time it queries the event detector; unnecessary queries may lead to a misunderstanding of task progress, leading to an increasingly high probability of task failure.
On the other hand, since querying the event detector is essential for the agent to know when the task is accomplished, controlling the query to the event detector is required to handle this trade-off.

We propose an RL framework, Learning Action and Query over Belief LTL (LAQBL), to learn an agent that can consider the diversity of LTL instructions due to uncertain event detection while avoiding task failure due to the unnecessary event-detection query. 
Our framework simultaneously learns three components: 1) \textit{embeddings of belief LTL} that embed belief LTL, a probabilistic representation of multiple LTL instructions caused by the uncertainty in the event detector, 2) \textit{action policy}, and 3) \textit{query policy} which decides whether or not to query for the event detector, with that embedding as input. 
We exploit LTL progression for task transition and the belief LTL is embedded by a graph neural network (GNN) \cite{Vaezipoor2021-qm} to obtain non-myopic policy and generalisability to unknown LTL formulas.
Our method is evaluated on the navigation in a grid world and the pipe inspection with high-dimensional image inputs under LTL instructions that require the agent to branch its behavior according to the subtask it achieves.
Experimental results on long-horizon tasks that require action according to event detection uncertainties show that our agent outperforms a method without handling the uncertainty and achieves the best performance even when unnecessary queries to the event detectors lead to task failure. 

Our main contributions are summarized as follows:
\begin{itemize}
    \item Formulation for optimizing an agent that acts according to LTL instructions with uncertain event detectors; 
    \item Proposal of a model for embedding multiple LTL instructions with their belief as belief LTL, inspired by LTL embedding in \cite{Vaezipoor2021-qm};
    \item Proposal of the LAQBL framework that learns the belief LTL embeddings, action, and query policies with the embeddings as input by reinforcement learning; 
    \item Empirical evaluations in navigation with a 2D grid-world and image-input robotic inspection simulation environments. 
\end{itemize}
\vspace{-2mm}

\section{RELATED WORKS}
\subsection{Multiple LTL Instructions for RL Agent}
To our knowledge, there are two frameworks for instructing various LTL instructions to an agent: Reward Machines (RM) \cite{icarte2018using,camacho2019ltl}, and LTL embedding \cite{leon2020systematic,kuo2020encoding,Vaezipoor2021-qm}.
RM provides automata representation constructed from symbolic task plans and enables designing the reward function, including non-Markovian rewards.
While RM has the flexibility to represent symbolic tasks implemented by various formal languages, generalizing to unknown tasks is difficult because no information about the task structure is embedded in its state.
The LTL embedding gives the agent a state that embeds part or the entire LTL instruction instead of defining explicit task states.
This approach successfully generalizes to unknown LTL instructions by learning the embedding function of LTL formulas.
However, most of these methods assume that the agent can detect the event that occurred in the environment and accurately track task progress.

\subsection{Planning under Uncertain LTL Instructions}
Previous works have considered addressing various uncertain situations encountered when implementing symbolic task representations in the real world to mitigate this assumption \cite{Guo2018,hasanbeig2019reinforcement, cai2020receding, Kantaros2020, cai2021modular}.
Particularly papers dealing with uncertainty regarding event detection are related to our proposal \cite{Sharan2014, bouton2020point, hashimoto2021collaborative, Gao2020, shah2020planning, li2022noisy}.
The method proposed in \cite{shah2020planning} is similar to our method of modeling a belief over multiple LTL formulas for learning through RL. 
The belief is used to define reward functions for four different objectives in this method, while it is not assumed that a given belief will be updated.
The main difference between ours and all these methods is that they do not provide a method to capture the uncertainty in the event detector; they cannot learn to act according to it.
Table \ref{tab:uncertain_method} summarizes the differences in belief models between our method and conventional.
\begin{table}[h]
\vspace{2mm}
    \begin{center}
    \begin{tabular}{c|cc}
    \hline
    Methods & Support of belief & Belief update trigger \\
    \hline 
    Sharan et al.\cite{Sharan2014} & States & No update \\
    Bouton et al.\cite{bouton2020point} & States & Observation \\
    Hashimoto et al.\cite{hashimoto2021collaborative} & States & Event detection\\
    RMSM\cite{li2022noisy} & Automaton states & Direct learning \\
    PUnS\cite{shah2020planning} & LTL formulas & No update \\
    Ours & LTL formulas & Event detection\\
    \hline
    \end{tabular}
    \vspace{-1mm}
    \caption{A comparison of modeling belief and its update trigger.}
    \label{tab:uncertain_method}
    \end{center}
\vspace{-6mm}
\end{table}
\vspace{-3mm}
\section{PRELIMINARIES}
\subsection{Linear Temporal Logic}
LTL formula consists of a finite number of propositional symbols $\mathcal{P}$, boolean operators to connect propositions such as $\land$ (conjunction), $\lor$ (disjunction), $\lnot$ (negation), and temporal operators to express the order of propositions such as $\cup$ (until), $\bigcirc$ (next), $\Diamond$ (eventually).
For instance, the sequence ``go to the kitchen and then go to the bedroom'' can be described as $\varphi=\Diamond (\mathtt{Kitchen}\land \Diamond \mathtt{Bedroom} )$.
In this paper, we consider co-safe LTL (sc-LTL) \cite{kupferman2001model}, a subclass of LTL that deals with sequences of finite length, and describe sc-LTL simply as LTL.
Detailed LTL syntax can be found in \cite{baier2008principles}.

\subsection{LTL-guided Policy for Taskable MDP}
In common RL, the interaction of an agent with the environment is modeled by a Markov Decision Process (MDP) and defined by the following tuple $\mathcal{M}=( S, T, A, p, R, \gamma, \mu)$, where $S$ is a set of states, $T\subseteq S$ is a set of terminal states, $A$ is a set of actions,  $P : S \times A \times S \rightarrow [0,1]$ is the state transition function, $R : S \times A \rightarrow \mathbb{R}$ is a reward function, $\gamma \in(0,1)$ is a discount factor, $\mu : S\to[0,1]$ is a distribution over initial states.

To obtain the agent that satisfies a given LTL instruction $\varphi$ through RL, the LTL task's progress and the environmental state must be managed simultaneously at every time step.
We define a word $\sigma_t\in\{0,1\}^{\mathcal{P}}$, a vector of propositions satisfied at time step $t$, and introduce a labeling function $L:S\rightarrow \{0,1\}^{\mathcal{P}}$ that maps a state to the word $\sigma_t$.
For RL with the LTL objective, the agent can be motivated to follow the LTL instruction $\varphi$ by rewarding that $\varphi$ holds in the finite trace $\sigma=(\sigma_0,\sigma_1,\ldots,\sigma_n)$, which is the mappings of the generated trajectory by the agent: $s_0,a_0,\cdots,s_n,a_n$.
However, the reward function $R_{\varphi}(s_1,a_1,\ldots,s_t,a_t)$ is conditioned on a past trajectory and actions and becomes non-Markovian \cite{toro2018teaching}.

In recent works, the LTL progression is used to make the reward function Markovian \cite{toro2018teaching, Vaezipoor2021-qm}, which implements the syntactic progression rule (see details in \cite{bacchus2000using}).
In brief, the LTL progression diminishes an LTL formula by leaving propositions unsatisfied while preserving the original semantics and returns $\mathsf{true}$ if the LTL formula is completed and $\mathsf{false}$ if it is violated.
For example, when an agent has the LTL instruction $\varphi=\lnot\mathtt{a}\cup\mathtt{b}$ at time step $t$, $\varphi$ is overwritten $\varphi:=\operatorname{prog}(L(s_{t+1}), \varphi)=\mathtt{true}$ by the progression if the event $\mathtt{b}$ is detected at $t+1$, $\varphi:=\mathtt{false}$ if the event $\mathtt{a}$ is detected, and does not change if any other event is detected.
\textit{Taskable MDP} proposed in \cite{Vaezipoor2021-qm} is a model that allows the agent to follow various LTL instructions while preserving Markov property by using the LTL progression and is defined as follows.
\begin{dfn}[\textit{Taskable MDP}\cite{Vaezipoor2021-qm}]
  Given a MDP without a reward function as a tuple $\mathcal{M}=\left( S, T, A, p, \gamma, \mu \right)$, a finite set of propositions $\mathcal{P}$, a labeling function $L$, a finite set of LTL formulas $\Phi$ and its probability distribution $\tau$ over $\Phi$, a Taskable MDP is defined as a tuple $\mathcal{M}_{\Phi}=\left( S^{\prime}, T^{\prime}, A, p^{\prime}, R^{\prime}, \gamma, \mu^{\prime}\right)$,  where $S^{\prime}=S\times cl(\Phi)$ is a finite set of product states and $cl(\Phi)$ is the smallest set containing $\Phi$ that consists of $\varphi\in\Phi$ and its progression.
  $T'=\{\langle s,\varphi\rangle|s\in T' \ \text{or} \  \varphi\in\{\mathsf{true,false}\}\}$ is a terminal set of product states, $p^{\prime}$ is a transition probability of product states; $p^{\prime}\left(\langle s^{\prime}, \varphi^{\prime}\rangle \mid\langle s, \varphi\rangle, a\right)=p\left(s^{\prime} \mid s, a\right)$ if $\varphi^{\prime}=\operatorname{prog}(L(s), \varphi)$ otherwise zero, $\mu^{\prime}(\langle s, \varphi\rangle)=\mu(s) \cdot \tau(\varphi)$ is an initial distribution of product states, and a reward function 
\begin{equation}
R^{\prime}((s, \varphi), a)= \begin{cases}1 & \text { if } \operatorname{prog}(L(s), \varphi)=\mathsf { true } \\ -1 & \text { if } \operatorname{prog}(L(s), \varphi)=\mathsf { false } \\ 0 & \text { otherwise }\end{cases}.
\label{eq:reward_ltl2action}
\end{equation}
\end{dfn}
\noindent
The LTL formula is mapped to the LTL feature space $\mathcal{E}\subseteq\mathbb{R}^n$ through the embedding $E_{LTL}:cl(\Phi)\rightarrow\mathcal{E}$, which is then given to the policy $\pi:S\times\mathcal{E}\times A\rightarrow [0,1]$.

\section{Problem formulation}
\begin{figure*}[t]
    \centering
    \vspace{2mm}  
    \includegraphics[width=168mm]{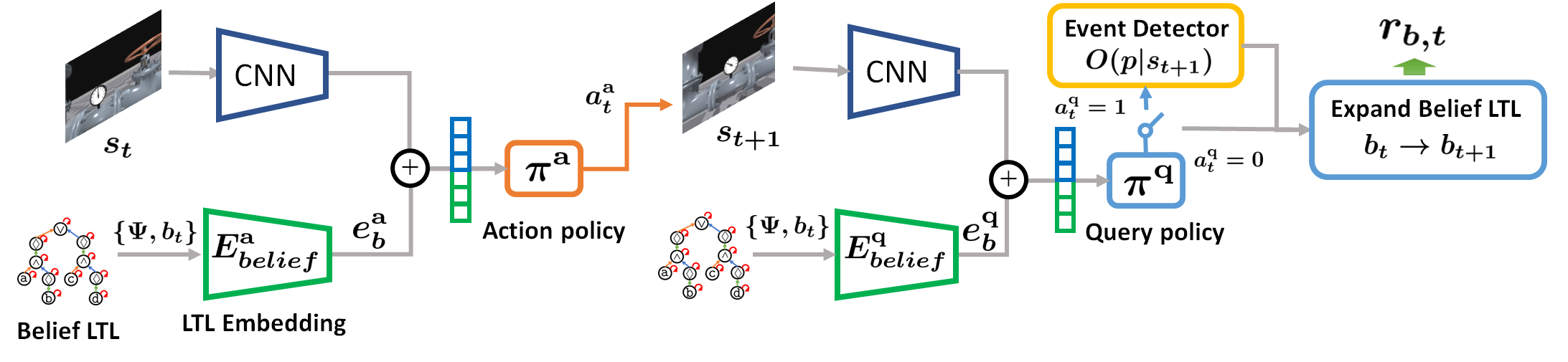}
    \vspace{-2mm}
    \caption{Overview of the LAQBL framework. The state $s$ and the belief LTL embeddings $e_b$ are input to the policies, and each action is sampled sequentially. All the embeddings and policies are optimized through RL by the reward $r_{b}$.}
    \label{fig:overview_proposed_encode}
    \vspace{-6mm}
\end{figure*}

\subsection{Modeling Event Detector and Belief over LTL formulas}\label{Belief_LTL}
We first replace a labeling function $L$, which deterministically provides the agent with the true events depending on the state, with a probabilistic event detector with uncertainty.
In this paper, we model the uncertainty of the event detector as a function that returns probabilities for the agent satisfying events for a given state as follows.
\begin{dfn}[\textit{Uncertainty of the event detector}]
Given a state $s_t$ in time step $t$, a finite set of propositions $\mathcal{P}$ and an LTL formula $\varphi$, the uncertainty of event detector is a probability distribution with support over $\mathcal{P}^{\prime}=\mathcal{P}\cup\lbrace\rbrace$, where $\{\}$ denote a null proposition, with the probability mass function $O:S\times\{ 0,1\}^{\mathcal{P}^{\prime}} \rightarrow [0,1]$, where $O(p|s_t)$ represents the confidence of the proposition that $p$ holds $\mathsf{true}$ in a state $s_t$ and $\sum_{p \in\mathcal{P}^{\prime}} O(p|s_t)=1, \forall s \in S$.
\end{dfn}
Then, we define \textit{possible LTLs}, a set of LTL formulas that can progress from the original LTL instruction by events for which the event detector has uncertainty, and a belief of possible LTL, as follows.
\begin{dfn}[\textit{Possible LTLs and a belief of possible LTLs}]
 Given a state $s_t$ in time step $t$ and the uncertainty of event detector $O(p|s_t)$, the possible LTLs is defined as a finite set of LTL formulas $\Psi=\bigcup_{p^o\in \mathcal{P}^o}\{\operatorname{prog}(\sigma^{p^o},\varphi)\}$ which consists of $\varphi\in\Phi$ and its progressed only propositions with positive probability $\mathcal{P}^o=\{p|O(p|s_t)>0, p\in\mathcal{P}^{\prime}\}$.
 A belief of the possible LTLs is defined as a probability distribution with support over the possible LTLs $B=\{b:\Psi\rightarrow[0,1]\mid\Psi\subset cl(\Phi),\sum_{\psi\in\Psi} b(\psi)=1\}$.
The expansion of the belief in the next time step $s_{t+1}$ is described as follows:
\begin{multline}
b_{t+1}(\psi^{\prime})= \\
\sum_{p^o\in\mathcal{P}^o,\psi\in\Psi} O(p^o|s_{t+1})b_{t}(\psi)\mathds{1}[\operatorname{prog}(\sigma^{p^o}, \psi)=\psi'].
\label{eq:belief_update}
\end{multline}
\end{dfn}
\noindent
\textbf{Example of updating the belief of possible LTLs.}\label{ex} Given a LTL formula: $\varphi=\Diamond (\mathtt{a}\land \Diamond \mathtt{b})\lor\Diamond (\mathtt{c}\land \Diamond \mathtt{d})$, which means ``go to $\mathtt{a}$ and then $\mathtt{b}$, or go to $\mathtt{c}$ and then $\mathtt{d}$''.
Assume that the belief $b_t(\varphi)=1$ and the uncertainty of the event detector in state $s_{t+1}$ are $O(\mathtt{a}|s_{t+1})=0.8$, $O(\mathtt{c}|s_{t+1})=0.2$ and otherwise 0. 
The propositions that may have probability are $\mathtt{a}$ and $\mathtt{c}$, and the possible LTLs are progressed by $P^o$: $\Psi=\{\varphi_\mathtt{a}=\operatorname{prog}(\sigma^{\mathtt{a}}, \varphi)=\Diamond \mathtt{b}\lor\Diamond(\mathtt{c}\land\Diamond\mathtt{d}),\varphi_\mathtt{c}=\operatorname{prog}(\sigma^{\mathtt{c}}, \varphi)=\Diamond (\mathtt{a}\land \Diamond \mathtt{b})\lor\Diamond\mathtt{d}\}$.
Then, the belief of the possible LTLs is updated by using their probabilities $b_{t+1}(\varphi_{\mathtt{a}})=0.8,b_{t+1}(\varphi_{\mathtt{c}})=0.2$.

\subsection{Belief Taskable MDP}
To model the process of the robot interacting with the environment and the event detector based on the belief, we introduce a query policy $\pi^\text{q}:S\times \mathcal{E}_b\times A_{\text{q}}\rightarrow [0,1]$ that optimizes the query timing to the detector in addition to the action policy $\pi^\text{a}:S\times\mathcal{E}_b\times A_{\text{a}}\rightarrow [0,1]$, where $\mathcal{E}_b$ is the feature space of the possible LTLs and its belief mapped by the embedding $E_{Belief}:\Psi\times B\rightarrow\mathcal{E}_b$. 
Here $A_{\text{q}}=\{0,1\}$, and we assume that the agent can only get uncertainty about propositions from the event detector when the query action sampled by the query policy is $a_{\text{q}}=1$, while the event detector returns a null proposition $O(\{\}|s)=1$ when the query action is $a_{\text{q}}=0$.
Finally, we construct a Belief Taskable MDP (BTMDP) as follows.
\begin{dfn}[\textit{BTMDP}]
Given a finite set of LTL formulas $\Phi$ and its probability distribution $\tau$ over $\Phi$, a finite set of propositions $\mathcal{P}$, a labeling function $L$, an event detector $O$ and a MDP without a reward function as a tuple $\mathcal{M}=\left( S, T, A, p, \gamma, \mu \right)$, the BTMDP is defined as $\mathcal{M}_{b}=\left( S_b, T, A_b, p_{b}, R_{b}, \gamma, \mu_{b}\right)$, where $S_{b}=S\times \Psi$ is a finite set of product states, $A_b=A_{\text{a}}\times A_{\text{q}}$ is a finite set of product action, $p_{b}\left(\left( s^{\prime}, b^{\prime}(\psi^{\prime})\right) \mid( s, b(\psi)), a\right)=p\left(s^{\prime} \mid s, a\right)\cdot b'(\psi')$, $\mu_{b}(s, b(\varphi))=\mu(s) \cdot \tau(\varphi)$ is a initial distribution of product states, where $\tau(\varphi)$ is the probability of selecting $\varphi$ from $\Phi$ and initially $b(\varphi)=1$, and a reward function 
\begin{equation}
\label{eq:reward_function_HLMDP}
 R_{b}((s, \Psi), a)=
  \begin{cases}
    1 & \text { if } \Psi\cap\{\psi_{truth}\}=\mathsf{true} \\
   0 & \text { otherwise }
  \end{cases}.
\end{equation}
\end{dfn}
\noindent
Here $\psi_{truth}$ is the LTL formula progressed by events captured by the labeling function, which the agent cannot directly observe, and used only for reward calculation during training.
The possible LTLs $\Psi$ manage all progressions of branched LTL formulas according to the uncertainty, which means that the progress of the LTL formula is partially observable for the agent.
Therefore, the policies for $\mathcal{M}_b$ can act following the belief of the possible LTLs but cannot know whether its actions follow the actual events in the environment during training.
The reward function $R_{b}$ addresses this situation by encouraging both $\psi\in\Psi$ and $\psi_{truth}$ to become $\mathsf{true}$ simultaneously.
On the contrary, if both $\psi\in\Psi$ and $\psi_{truth}$ are unsatisfied or progress to $\mathsf{false}$ either of these, the agent is not rewarded.

\noindent
\textbf{Example of reward calculation.} Consider the example in Section \ref{ex}: $\Psi=\{\varphi_\mathtt{a}=\Diamond \mathtt{b}\lor\Diamond (\mathtt{c}\land \Diamond \mathtt{d}),\varphi_\mathtt{c}=\Diamond (\mathtt{a}\land \Diamond \mathtt{b})\lor\Diamond\mathtt{d}\}$.
At the next time step $t+1$ if $\varphi_{truth}=\varphi_\mathtt{a}$:
\begin{itemize}
    \item when $L(s_{t+1})=\mathtt{b}$ and the agent receives $O(\mathtt{b}|s_{t+1})=1.0$, $\operatorname{prog}(\sigma^{\mathtt{b}}, \varphi_{\mathtt{a}})=\operatorname{prog}(\sigma^{\mathtt{b}}, \varphi_{truth})=\mathsf{true}$ and the agent obtains a reward of 1.
    \item when $L(s_{t+1})=\mathtt{d}$ and the agent receives $O(\mathtt{d}|s_{t+1}))=1.0$, only $\operatorname{prog}(\sigma^{\mathtt{d}}, \varphi_{\mathtt{c}})=\mathsf{true}$ and the agent obtains a reward of 0.
\end{itemize}
Note that to make a reward calculation of an episode feasible, we assume that the event detection of propositions regarding the terminal set is done with no uncertainty in this paper.

\subsection{Problem Statement}
This paper considers the problem of an agent interacting with an environment and an event detector modeled by a BTMDP.
Our goal is to obtain policies that maximize the reward obtained by correctly progressing the instructed LTL under the belief state based on the uncertain detection of propositions obtained from the event detector.
To this end, the problem can be described as follows.
\begin{plm}
Given a BTMDP $\mathcal{M}_{b}=\left( S_{b}, T_{b}, A_b, p_{b}, R_{b}, \gamma, \mu_{b}\right)$, a finite set of propositions $\mathcal{P}$, a labeling function $L$, a finite set of LTL formulas $\Phi$, a probability distribution $\tau$ over $\Phi$ and an event detector $O$, find policies $\pi^{\text{a}}_b$ and $\pi^{\text{q}}_b$ that optimize all tasks in the set $\Phi$ by maximizing the reward $R_{b}$ for all state $s_b\in S_b$ and LTL instructions $\varphi\in\Phi$.
\end{plm}

\section{Learning Action and Query policies Over Belief LTL}
\subsection{Algorithm Overview}
We propose a framework, namely Learning Action and Query over Belief LTL (LAQBL), that simultaneously learns the belief LTL embedding $E_{belief}$, which learns the representation of belief LTL, and two policies, the action policy $\pi^{\text{a}}$ and the query policy $\pi^{\text{q}}$, all through RL.
We use different functions $E^{\text{a}}_{belief}$ and $E^{\text{q}}_{belief}$ of the same architecture for each policy $\pi^{\text{a}}$ and $\pi^{\text{q}}$ with different objectives.
An overview of the LAQBL is illustrated in Fig \ref{fig:overview_proposed_encode}.

Given an LTL formula $\varphi$ sampled from the set $\Phi$ according to a distribution $\tau$ and an initial state $s_0$, actions from the two policies are sequentially sampled.
Firstly, the state $s_0$ and the feature of the belief LTL  $e^{\text{a}}_b=E^{\text{a}}_{belief}(\varphi,b(\varphi)),e^{\text{a}}_b\in\mathcal{E}_b,b(\varphi)=1.0$ are input to the policy $\pi^{\text{a}}(a^{\text{a}}|s_0,e^{\text{a}}_{b,0})$ and an action $a^{\text{a}}_0$ is sampled. 
Next, the state $s_1$ transitioned by the state transition probability $p_{b}=p(s_1|s_0,a^\text{a}_0)$ is input to the policy $\pi^{\text{q}}(a^{\text{q}}|s_1,e^{\text{q}}_{b,0})$.
The possible LTLs $\Psi$ are computed from LTL formulas progressed by propositions with $O(p|s_1)$ greater than 0, and the belief LTL is updated by Eq. (\ref{eq:belief_update}).
Finally, the state $s_1$ and the belief LTL again input to the policy $\pi^{\text{a}}$.
Both policies are optimized using the reward $r_{b}$ through RL until the end of the episode.

\vspace{-1mm}
\subsection{Graph Embedding of Belief LTL}\label{Encoding_Belief}
For the agent to learn behavior based on various given LTL instructions while considering the uncertainty of event detectors, it is crucial for the belief LTL embedding $E_{belief}$ to learn discriminative features based on the differences in LTL instructions and their beliefs.
A GNN-based method that embeds including future events \cite{Vaezipoor2021-qm} leads to non-myopic behavior and generalizability to unseen LTL formulas compared to a naive method that encodes only the next event symbols in given LTL formula \cite{leon2020systematic} or RM-based method \cite{camacho2019ltl}.
However, these methods do not provide a way of capturing uncertainty in event detection.
To tackle this issue, we propose an augmentation of the GNN-based embedding that preserves LTL semantics and embeds both branched LTL formulas and their beliefs into the feature space.

We first represent the LTL formulas in the possible LTLs $\psi\in\Psi$ as a tree-structured directed graph with the tokens (propositions and operators) assigned to nodes in the same way as in \cite{Vaezipoor2021-qm}.
As an example, the graph representation of the possible LTLs $\Psi=\{\varphi_\mathtt{a}=\Diamond \mathtt{b}\lor\Diamond (\mathtt{c}\land \Diamond \mathtt{d}),\varphi_\mathtt{c}=\Diamond (\mathtt{a}\land \Diamond \mathtt{b})\lor\Diamond\mathtt{d}\}$ is shown in Fig. \ref{fig:belief_LTL} (a).
We add a new node connected to the generated graphs and assign the beliefs $b(\varphi_{\text{a}})$ and $b(\varphi_{\text{c}})$ corresponding to each LTL formula as the weight of the connected edges $W_{\varphi_\text{a}}$ and $W_{\varphi_\text{c}}$.
We define the constructed LTL graph as a \textit{belief LTL}, representing the set of the possible LTLs and their belief.
The embedding of the belief LTL $e_b$ by the following formula, with a slight abuse of notation:
\begin{eqnarray}
e_b &= E_{belief}(\{\psi,b(\psi)|\psi\in\Psi,b(\psi)\in B\})\\ \nonumber
&=f\left(\sum_{\psi\in\Psi}x_{\psi}W_\psi\right),
\end{eqnarray}
where $f$ is a readout function, we use a 1-layer fully-connected network, and $x_\psi$ is the node feature of the LTL graph $\psi\in\Psi$ by relational graph convolutional network (R-GCN) \cite{schlichtkrull2018modeling}.
Since message passing by R-GCN is performed bottom-up, the feature $x_\psi$ of the top node of each LTL graph is treated as the feature of the graph.
The belief LTL can build unique graph representations according to the uncertainty of the event detector even if given the same instructions and can scale according to the belief expansion by Eq. (\ref{eq:belief_update}).

\vspace{-1mm}
\section{Experiments}
\subsection{Evaluation of the Belief LTL Embeddings}\label{sec:eval_belief}
\begin{figure}[t]
    \begin{minipage}{1.0\hsize}
    \centering
    \includegraphics[width=79mm]{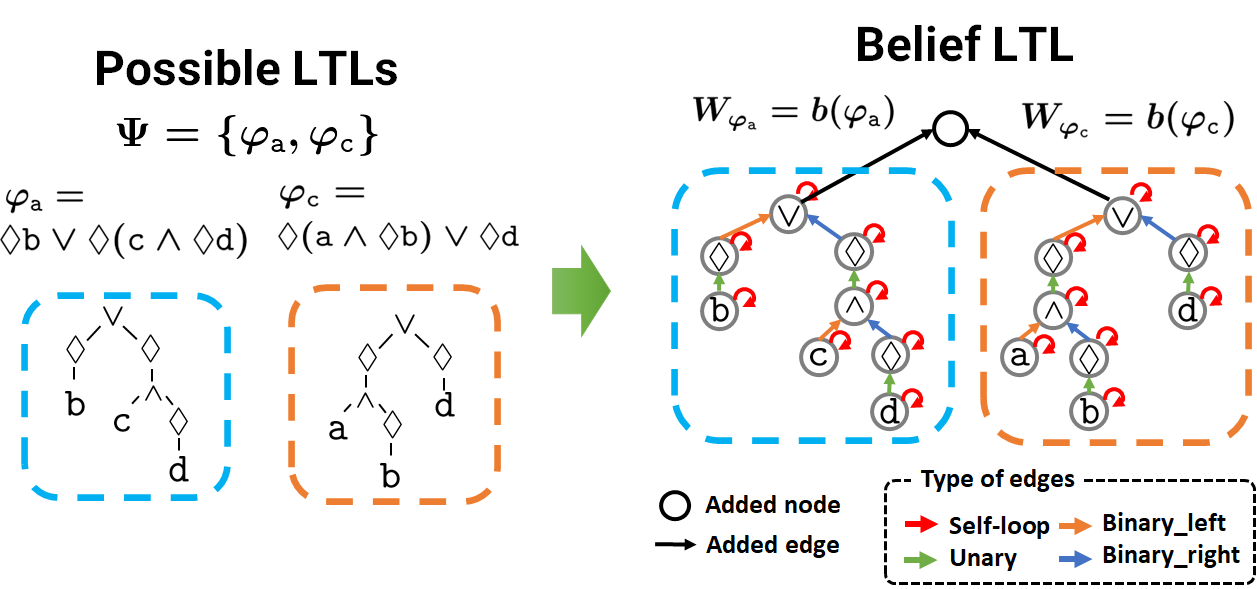}
    \vspace{-2mm}
    \subcaption{}
    \label{fig:belief_LTL_tree}
    \end{minipage}\\
    \begin{minipage}{1.0\hsize}
    \centering
    \includegraphics[width=85mm]{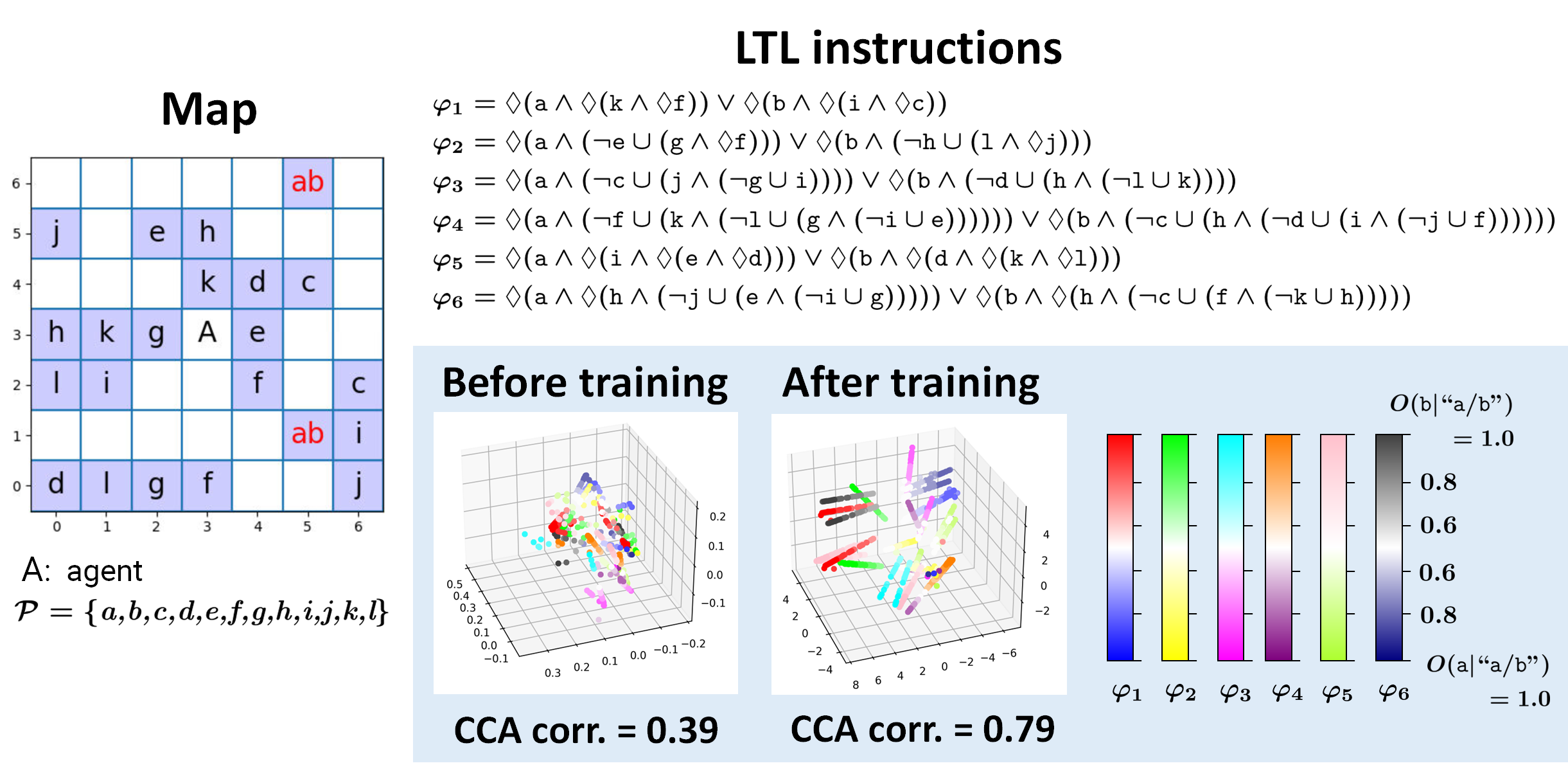} 
    \vspace{-1mm}
    \subcaption{}
    \label{fig:belief_embedding}
    \end{minipage}
    \vspace{-1mm}
   \caption{(a) A graph representation of the belief LTL. Each node in the graph is assigned a one-hot vector unique to the LTL token as a feature, and each edge type has a different weight. (b) Visualization of the belief LTL embeddings. The embeddings are represented in three-dimensional space by the PCA, with each color type and intensity indicating the given LTL instructions and its level of uncertainty. The correlation coefficient of the CCA is calculated by the features reduced by the PCA and the uncertainty observed in the ``$\mathtt{a}\mathtt{b}$" grid.}
    \label{fig:belief_LTL}
    \vspace{-5mm}
\end{figure}

\vspace{-1mm}
We first show whether the belief LTL embedding learns discriminative representation in the latent feature space according to given LTL formulas and their belief through a navigation task with a small LTL task set.
A map in Fig. \ref{fig:belief_LTL} (b) consists of 7$\times$7 squares with 10 unique events and the ``$\mathtt{a}\mathtt{b}$" placed twice.
As the state, each grid has 13 dimensions (12 events and the agent position), with 1 assigned to the channel corresponding to the elements present in each grid and 0 otherwise.
At the grid ``$\mathtt{a}\mathtt{b}$", the agent satisfies only one of the events in each episode and receives its uncertainty from the event detector.
The actions are moving up, down, left, and right.
The LTL task is uniformly sampled one of six tasks including until operator $\Phi=\{\varphi_1,\cdots,\varphi_6\}$ in Fig. \ref{fig:belief_LTL} (b).
The event detector can correctly detect the satisfied event with a random probability of 0.6 to 1.0 with a resolution of 0.01, and the agent observes this probability as its uncertainty in the grid ``$\mathtt{a}\mathtt{b}$".

We visualize the concatenated state vector and embedding of belief LTL through the rollout of the learned action policy by a principal component analysis (PCA).
In the qualitative evaluation, LTL instructions and their uncertainty level are organized in the feature space and represented discriminatively compared to before learning.
We also quantify the relation between features and belief using canonical correlation analysis (CCA) and confirm that our embeddings effectively capture the belief for each LTL instruction, as shown by the increase in canonical correlation after learning. 
\vspace{-5mm}
\subsection{Navigation under Different Uncertainty in Event Detector}
We then evaluate our method in the navigation that requires an agent to change actions depending on the event detectors' uncertainty level.
We use the same map as in the experiment in \ref{sec:eval_belief}.
One episode consists of 200 timesteps, and each $\pi^{\text{a}}$ and $\pi^{\text{q}}$ policy action consumes one.
Locations of events on the map are randomized for each seed, and the agent is always placed in the center. 
\subsubsection{LTL Task Design}\label{task_design}
\begin{figure}[t]
    \centering
    \vspace{2mm}  
    \includegraphics[width=75mm]{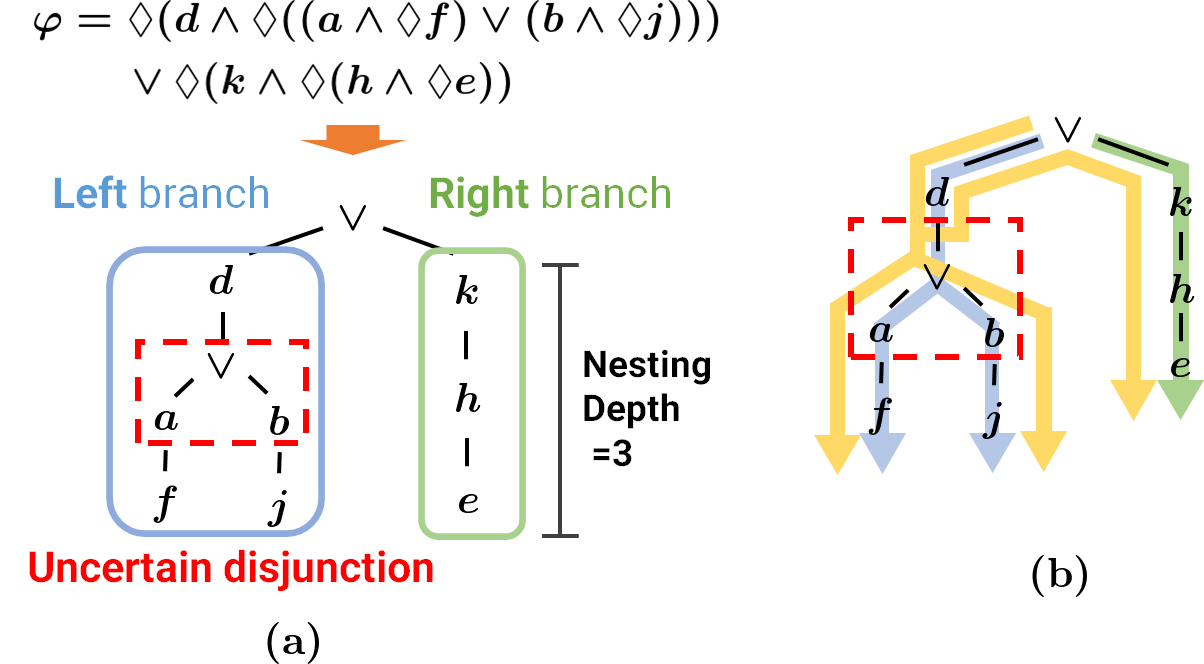} \\
    \centering\begin{subfigure}{1.0\linewidth}
    \includegraphics[width=81mm]{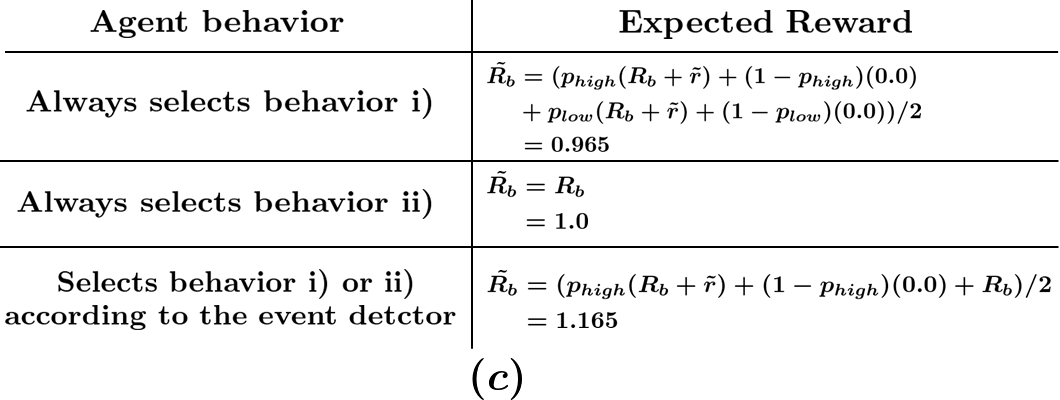}
    \vspace{-1mm}
    \end{subfigure}
    \caption{(a) An example of an LTL instruction given to an agent in the experiments (omits some operators for illustration). (b)(c) The reactive behavior of the agent on the LTL task tree and the corresponding expected returns that the agent obtains in the experiment. For the agent to obtain the highest return, it is necessary to choose a proposition to progress according to the event detector's uncertainty obtained in the ``$\mathtt{a}\mathtt{b}$" grid.}
    \label{fig:grid_overview}
    \vspace{-2mm}
\end{figure}

Unlike the common target focusing on whether the agent can accomplish LTL instructions, evaluating what traces the agent follows to accomplish LTL instructions is necessary for our objective.
Therefore, we design parallelized LTL task space to emphasize situations that motivate the agent to vary the action depending on the event detector's uncertainty.
Fig. \ref{fig:grid_overview} (a) shows an example of the LTL task used in our experiments.
The task is divided into a left and a right branch by the first disjunction operator ($\lor$), and the left branch has a disjunction corresponding to the ``$\mathtt{a}\mathtt{b}$" grid, making up the LTL task that can be satisfied by three different word traces.
We denoted the disjunction containing ``$\mathtt{a}\mathtt{b}$" as the \textit{uncertain disjunction} for convenience.

While this LTL task can be progressed to $\mathsf{true}$ no matter which traces the agent follows, we make it feasible for the agent to learn to choose these traces depending on the uncertainty of the event detector.
We use two types of event detector that probabilistically assigns uncertainty to the event: \textit{Expert} that detects the event with high accuracy $p_{high}$ and \textit{Beginner} that detects with low accuracy $p_{low}$.
This paper sets these to $p_{high}=0.95, p_{low}=0.5$, respectively, and fixed during training and testing.
That is, if the true event occurred on the grid ``$\mathtt{a}\mathtt{b}$" is $\mathtt{a}$, \textit{Expert} assigns $O(\mathtt{a}|s)=0.95$ and $O(\mathtt{b}|s)=0.05$ with 95\% and 5\% probabilities, respectively. On the other hand, \textit{Beginner} always assigns $O(\mathtt{a}|s)=0.5$ and $O(\mathtt{b}|s)=0.5$.
Since we do not inform the agent which detector is set and the event detectors are uniformly sampled at the start of the episode, the agent needs to select the following reactive behavior corresponding to the event detector's uncertainty.
\renewcommand{\theenumi}{\roman{enumi}}
\begin{enumerate}
    \item  If the agent observes low uncertainty at ``$\mathtt{a}\mathtt{b}$", it acts to achieve the event following the confident event (the trace indicated by blue on the LTL tree in Fig.  \ref{fig:grid_overview} (b)).
    \item If the agent observes high uncertainty at ``$\mathtt{a}\mathtt{b}$", it avoids the event following the uncertain disjunction with a high risk of failure (the trace indicated by green).
\end{enumerate}
\noindent
The most desirable way for the agent to address the randomness of the choice of event detectors during training is to change these behaviors according to the uncertainty obtained in the ``$\mathtt{a}\mathtt{b}$" grid, which corresponds to the trace indicated by yellow arrows on the LTL tree.

Note that we provide an additional reward $\tilde{r}$ if the agent progresses events through the trace containing the uncertain disjunction and earns a positive reward according to Eq. \eqref{eq:reward_function_HLMDP}.
This is necessary to balance the expected returns the agent can earn by behavior i) and ii), and we set the $\tilde{r}=0.4$ in this experiment.
The table in Fig. \ref{fig:grid_overview} (c) summarizes the expected returns obtained in the traces progressed by the agent.

\subsubsection{Training Details}
We use PPO \cite{schulman2017proximal} to train both policies $\pi^{\text{a}}$ and $\pi^{\text{q}}$, and the hyperparameters are the same as \cite{Vaezipoor2021-qm}.
The state is encoded by a 3-layer CNN and ReLU activations.
The LTL task $\varphi\in\Phi$ is uniformly sampled in each episode, and the set $\Phi$ has the uncertain disjunction in either branch, and propositions are randomized except for $\mathtt{a}$ and $\mathtt{b}$.
The nesting depth of the LTL formula is also set randomly from 2 to 4 (Fig. \ref{fig:grid_overview} (a) shows a tree with depth 3), and the number of possible unique tasks is over 197 million.
We focus on analyzing the agent behavior due to uncertainty in the ``$\mathtt{a}\mathtt{b}$" grid; the event detector returns the same results as the labeling function $L$ except for the ``$\mathtt{a}\mathtt{b}$" grid.
\subsubsection{Comparisons}
We evaluated our model against the following methods:
\begin{itemize}
    \item \textbf{LTL2Action:} The method employs the LTL embedding the same as the GNN-based method proposed in \cite{Vaezipoor2021-qm} and does not use the belief over multiple LTL formulas. The LTL formulas are progressed based on the Most Likely criteria \cite{shah2020planning}\footnote{The three reward functions except Most likely proposed in \cite{shah2020planning} are not in comparison because they aim to satisfy all LTLs that support the belief and are not consistent with the purpose of the experiment.}, which use a proposition with high probability to address uncertainty in the event detector.
    \item \textbf{belief+regular query:} The method uses the embedding of belief LTL the same as ours and performs a query to the event detector every step. This corresponds to the ablation of query policy in our method.
\end{itemize}
\subsubsection{Results}
\begin{figure*}[t]
    \vspace{2mm}  
    \begin{minipage}{0.24\hsize}
    \centering
    \includegraphics[width=40mm]{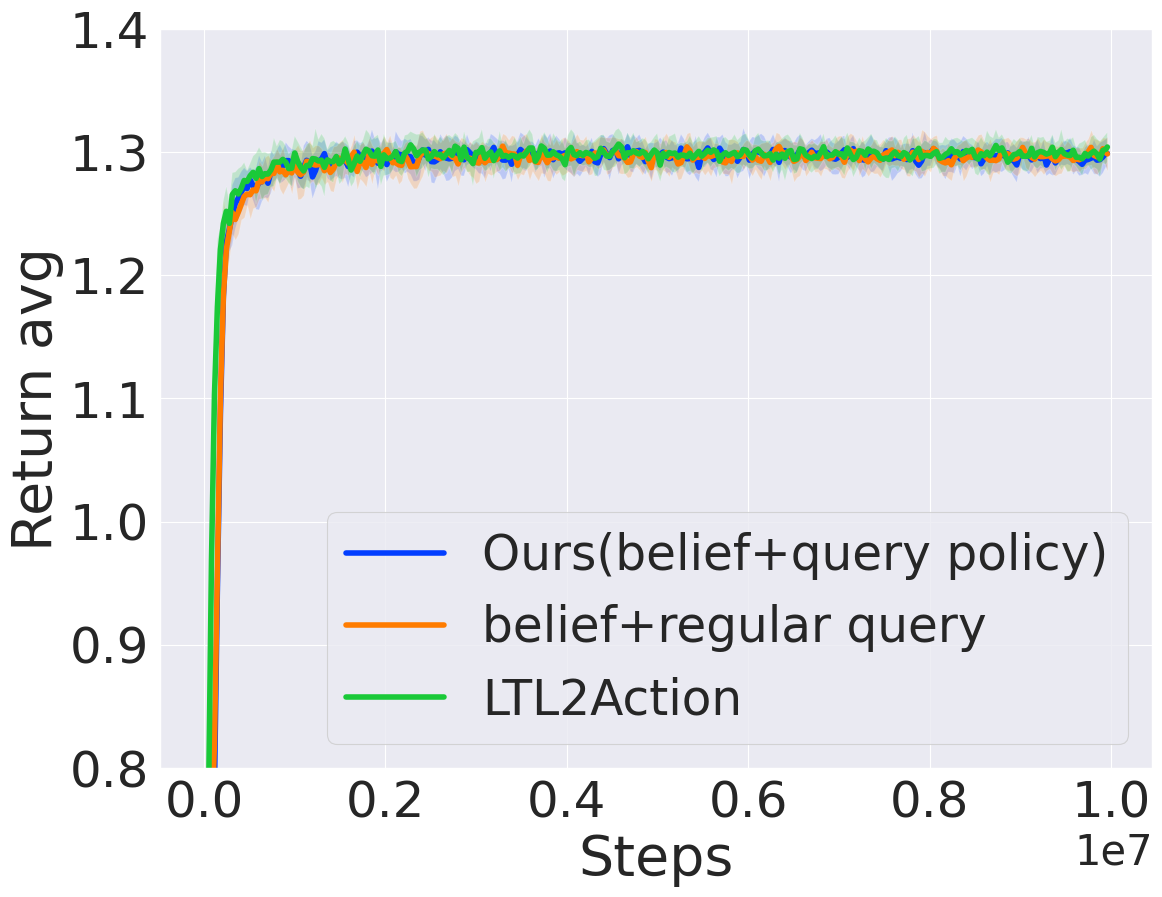}
    \vspace{-2mm}
    \subcaption{}
    \label{fig:grid_nopena_return}
    \end{minipage}
    \begin{minipage}{0.24\hsize}
    \centering
    \includegraphics[width=40mm]{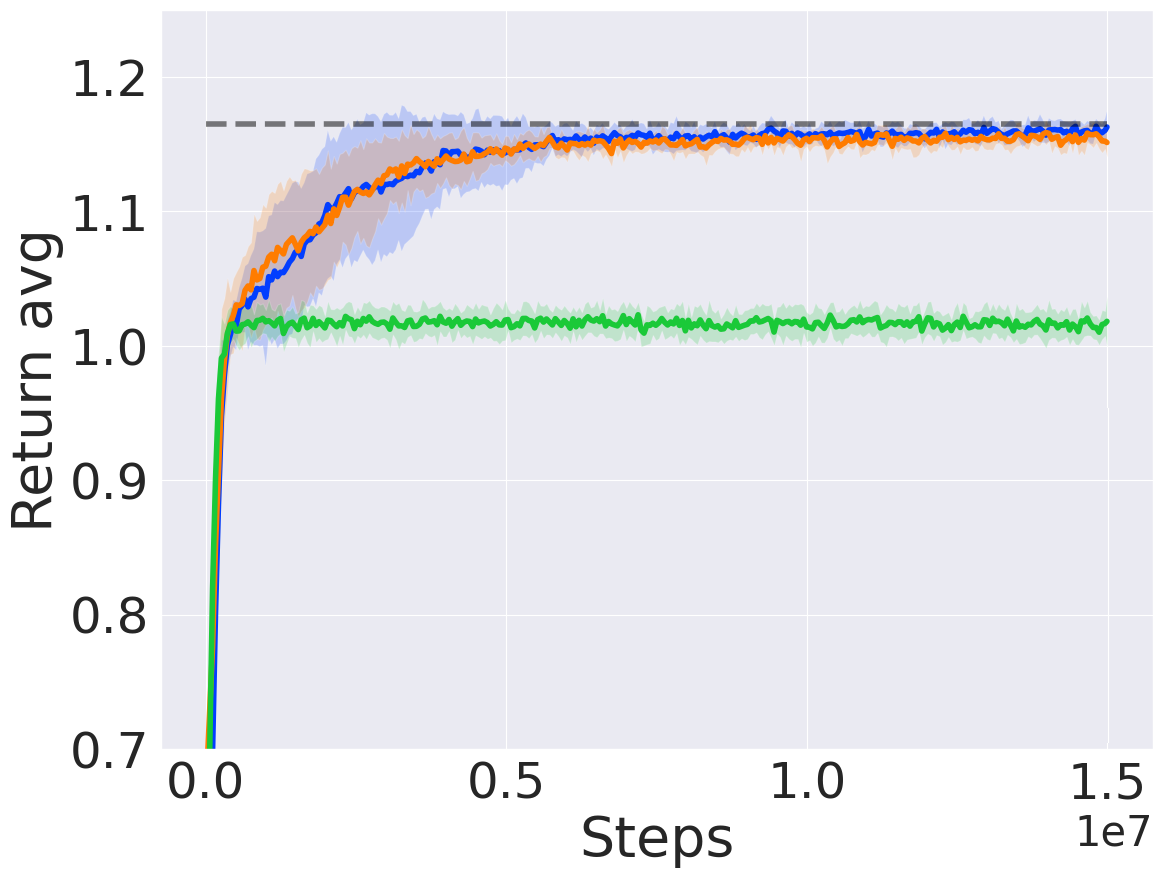}
    \vspace{-2mm}
    \subcaption{}
    \label{fig:grid_nopena_ratio}
    \end{minipage}
    \begin{minipage}{0.24\hsize}
    \centering
    \includegraphics[width=40mm]{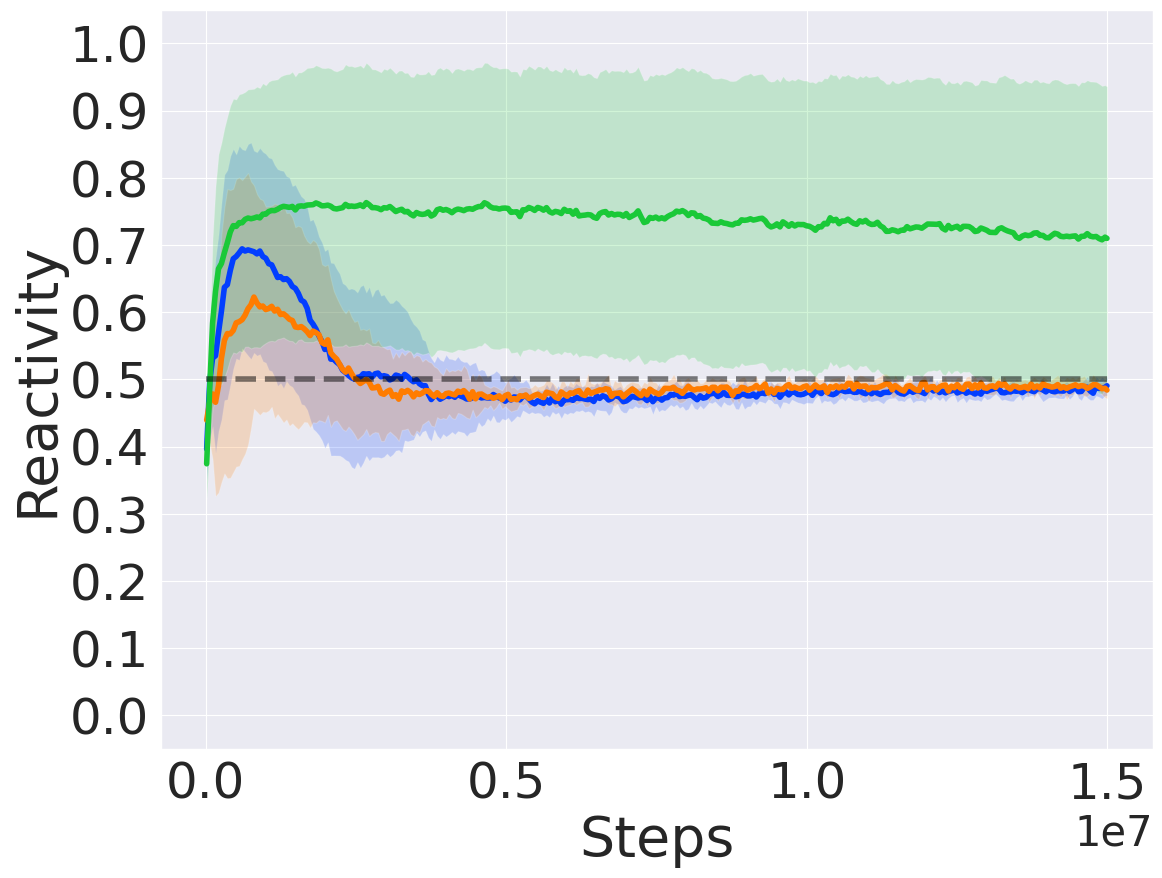}
    \vspace{-2mm}
    \subcaption{}
    \label{fig:grid_arch_return}
    \end{minipage}
    \begin{minipage}{0.24\hsize}
    \centering
    \includegraphics[width=40mm]{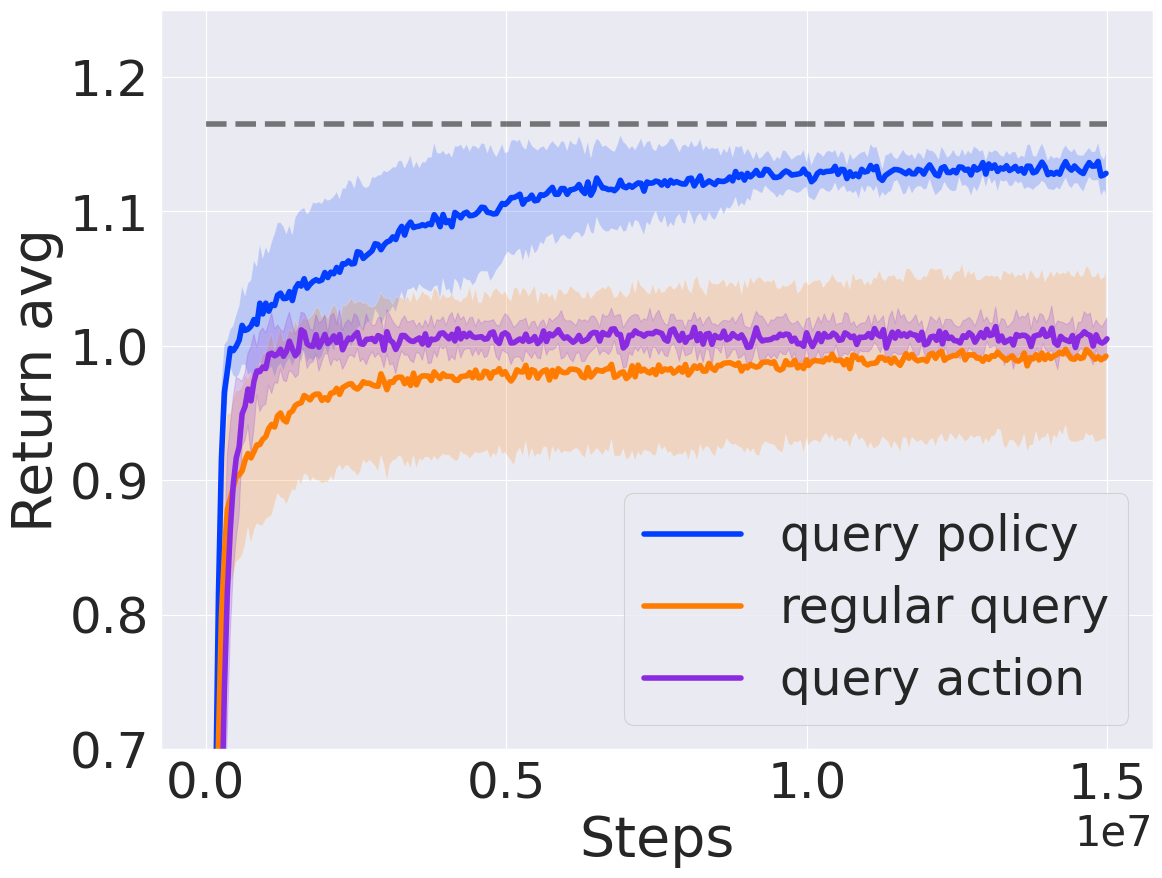}
    \vspace{-2mm}
    \subcaption{}
    \label{fig:grid_arch_frame}
    \end{minipage}
    \vspace{-2mm}
    \caption{(a)(b) The average returns with baselines during training in the navigation task with 15 random seeds. The probability that \textit{Expert} event detector is sampled is 95\% in (a) and 50\% in (b). (c) The reactivity of the agent to uncertainty in the event detector in the training of (b). Closer to 1.0 indicates that the agent achieves the task through the uncertain disjunction, and closer to 0.0 indicates that the agent achieves by avoiding that.  (d) The average returns with different policy architectures.}
    \label{fig:grid_result}
    \vspace{-7mm}
\end{figure*}

\noindent
\textbf{Evaluation of reactive behavior against different uncertainty in the event detector.} 
We first evaluate how different uncertainty in the event detector affects the behavior and performance of the agent.
Fig. \ref{fig:grid_result} (a) and (b) show the average return of each method during training for different probabilities of selecting \textit{Expert}  and \textit{Beginner} event detectors, respectively.
The probability of selecting \textit{Expert} sets to 95\% in Fig. \ref{fig:grid_result} (a) and 50\% in Fig. \ref{fig:grid_result} (b).
There is no difference in performance in the setting where \textit{Expert} is selected dominantly, whereas LTL2Action, which does not consider the belief, performs significantly worse in the setting where \textit{Expert} and \textit{Beginner} are uniformly selected.
Additionally, to quantify the reactivity of the agents to the uncertainty during training in the setting where the event detector is uniformly sampled, we count whether the agent performed the behavior i) and ii) in Section \ref{task_design} in Fig. \ref{fig:grid_result} (c).
The result that the reactivity of the belief-based method is closest to 0.5 and obtains the highest return in Fig. \ref{fig:grid_result} (b) indicates that the belief-based agents achieve select behavior i) or ii) according to the uncertainty in the event detector.
Alternatively, LTL2Action has a large variance of reactivity, indicating that it can not act according to the event detector.

\begin{table}[t]
\centering
\scalebox{0.93}{
\begin{tabular}{c|cccc}
\hline
& \multicolumn{4}{c}{Depth2-4}\\
Method & RT & RCT(\%) &  NEs & QFR(\%)\\ \hline
Ours(query policy) &\textbf{1.13}$\pm$0.30&\textbf{95.1}&9.27$\pm$17.52&2.6 \\
LTL2Action &  0.99$\pm$0.51&$49.1^{**}$&6.98$\pm$9.36&2.7  \\
belief+regular query & 0.99$\pm$0.46&$64.5^{**}$&0.67$\pm$0.93&-  \\
query action & 1.01$\pm$0.49&$49.6^{**}$&15.33$\pm$27.55&5.6\\ \hline\hline
& \multicolumn{4}{c}{Depth5}\\
Method & RT & RCT(\%) &  NEs & QFR(\%)\\ \hline
Ours(query policy) &\textbf{1.07}$\pm$0.37&\textbf{86.9}&18.04$\pm$24.09&3.1 \\
LTL2Action &  0.96$\pm$0.50&$48.3^{**}$&13.86$\pm$18.53&2.5  \\
belief+regular query & 0.91$\pm 0.53^{\dagger}$&$54.9^{**}$&1.2$\pm$1.59&-  \\
query action & 0.68$\pm$$0.66^{*}$&$33.1^{**}$&57.51$\pm$653.08&6.3\\ \hline
\end{tabular}
}
\vspace{0.2mm}
\\\leftline{$\dagger$: $\textit{p}<.05$, *: $\textit{p}<.005$, **: $\textit{p}<.0005$ by t-test.}
\caption{Results of testing policies in the navigation task under the probabilistic error of the event detector. We report the mean and standard deviation of the returns (RT), the mean accuracy of reactivity (RCT), the mean and standard deviation of the number of times to reach the grid with no events (NEs), and the query failure rate which averaged over queries on the grid with no events (QFR). All results are averaged over 15 seeds of 1000 episodes of the rollout of the learned policies, with 500 episodes of each fixed event detector per seed. The rollouts are performed on the training distribution of tasks (Depth2-4) and out-of-distribution tasks with increased depth of sequences (Depth5).}
\label{table:test_result}
\vspace{-6mm}
\end{table}
\vspace{-4mm}

\begin{figure}[t]
    \centering
    \includegraphics[width=82mm]{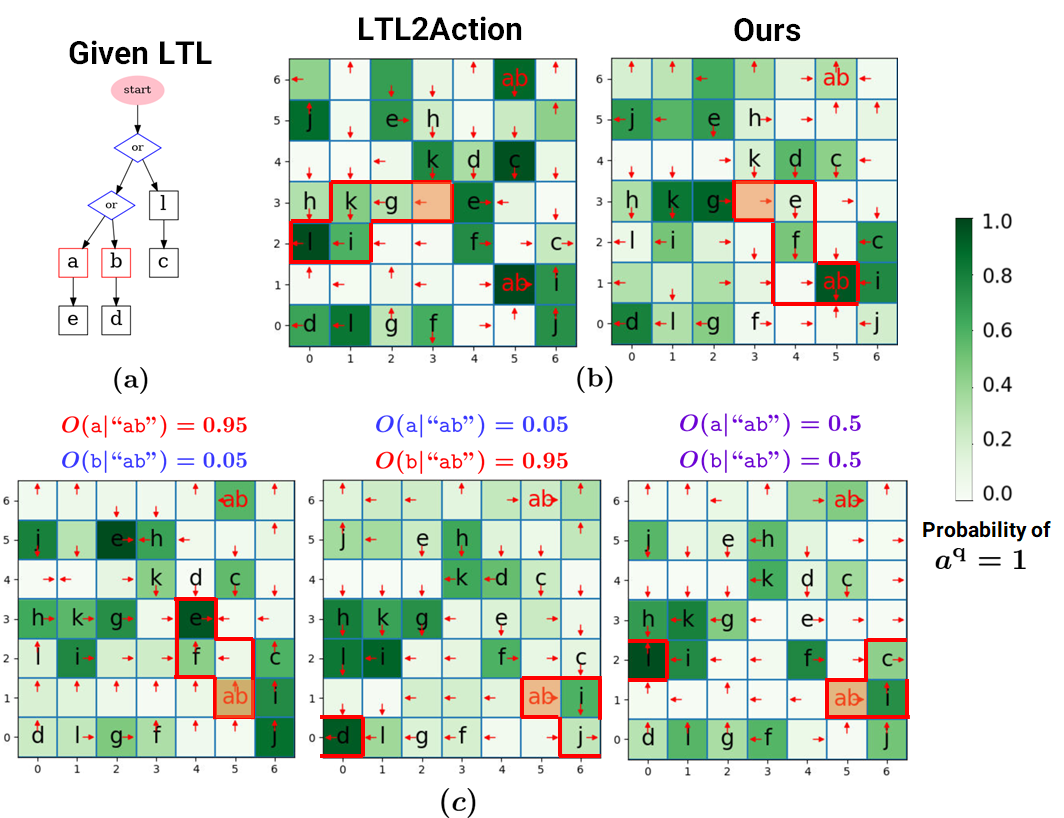}
    \vspace{-0.5mm}
    \caption{Visualisation of learned policies. (a) Flowchart of instructed LTL $\varphi=\Diamond(\mathtt{a}\land\Diamond\mathtt{e})\lor\Diamond(\mathtt{b}\land\Diamond\mathtt{d})\lor\Diamond(\mathtt{l}\land\Diamond\mathtt{c})$. (b) Comparison of actions of policies learned by LTL2Action and ours at the start of the episode. (c) Actions of our policies according to different uncertainties of the event detector after a query on the ``$\mathtt{a}\mathtt{b}$" grid at $(x,y)=(5,1)$. In each grid, red arrows indicate the dominant action of the policy $\pi^{\text{a}}$, and green intensity indicates the probability of action $a^{\text{q}}=1$ sampled by the policy $\pi^{\text{q}}$.}
    \label{fig:visualize_policies}
    \vspace{-5mm}
\end{figure}

Finally, we visualize the differences between the policies learned by LTL2Action and our method for qualitative evaluation in Fig. \ref{fig:visualize_policies}.
The rollout of learned policies with the LTL instruction in Fig. \ref{fig:visualize_policies} (a) at the map in Fig. \ref{fig:belief_LTL} (a) shows that LTL2Action ignores the ``$\mathtt{a}\mathtt{b}$" grid containing uncertainty, as shown in Fig. \ref{fig:visualize_policies} (b).
This result indicates that LTL2Action cannot capture the uncertainty and therefore learns only to achieve events not included in the uncertain disjunction.
On the other hand, our policy goes to the ``$\mathtt{a}\mathtt{b}$" grid first and achieves the events while following the instructions according to the uncertainty, as shown in Fig. \ref{fig:visualize_policies} (c).

\noindent
\textbf{Evaluation of the different query architecture under the probabilistic error of the event detector.} 
To validate the effect of the query policy of the LAQBL framework, we next experiment in the environment with probabilistic false positives by the event detector.
To directly implement the harmful effects of unnecessary queries, we set the task to fail immediately with a 10\% probability of failure if the policy $\pi^{\text{q}}$ queries the event detector on grids with no events in this experiment.
We compared the LAQBL framework with two methods, the regular query, which is an ablation study of our method, and the \textit{query action}, which adds a query action to the action policy $\pi^{\text{a}}$.
The result in Fig. \ref{fig:grid_result} (d) highlights the effectiveness of the LAQBL framework, and both the regular query and the query action performance are degraded.
Contrary to expectations, the regular query learns to move as far as possible through the grid where any events exist, but performance degradation is inevitable.
The query action lacks the independent query policy, making it hard to learn the embedding that optimizes both action and query policy and does not offer generalization performance.

Finally, we report the test result in Table \ref{table:test_result}.
Our method achieved the best performance and reactivity to uncertainty in the event detector for both LTL tasks in the training distribution and out-of-distribution with increased depth.
Since we are testing with a different random seed than in training, it is possible that LTL not sampled in training will be sampled in testing. The average returns of the test are comparable to the training result, which indicates that learned policies are optimized for the overall task set $\Phi$ as we expected.
Note that the query failure rate for the out-of-distribution task is slightly higher in our method, but this is because more moves and queries are required to accomplish longer tasks.
See the attached video of how our agent behaves.
\begin{figure}[t]
    \begin{subfigure}{.99\linewidth}
    \centering
    \includegraphics[width=80mm]{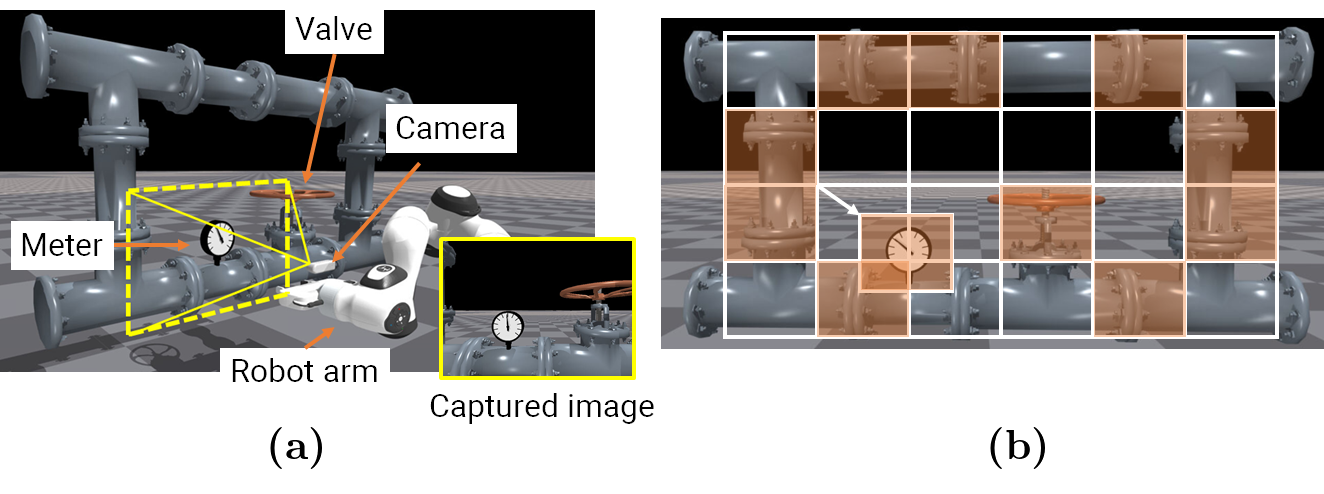}
    \end{subfigure}\\
    \vspace{-1mm}
    \begin{subfigure}{0.47\linewidth}
    \centering\setcounter{subfigure}{2}
    \includegraphics[width=40mm]{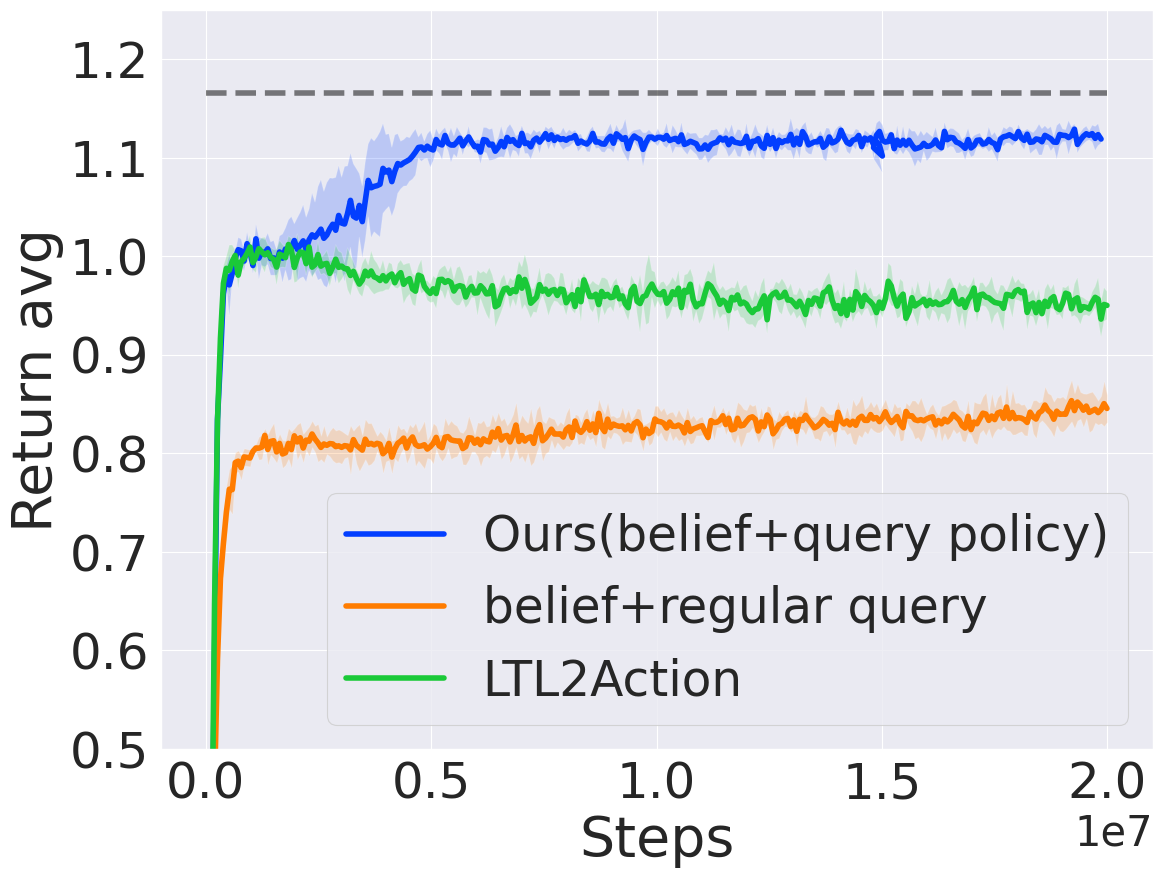}
    \vspace{-5mm}
    \subcaption{}
    \end{subfigure}
    \begin{subfigure}{0.47\linewidth}
    \centering
    \includegraphics[width=40mm]{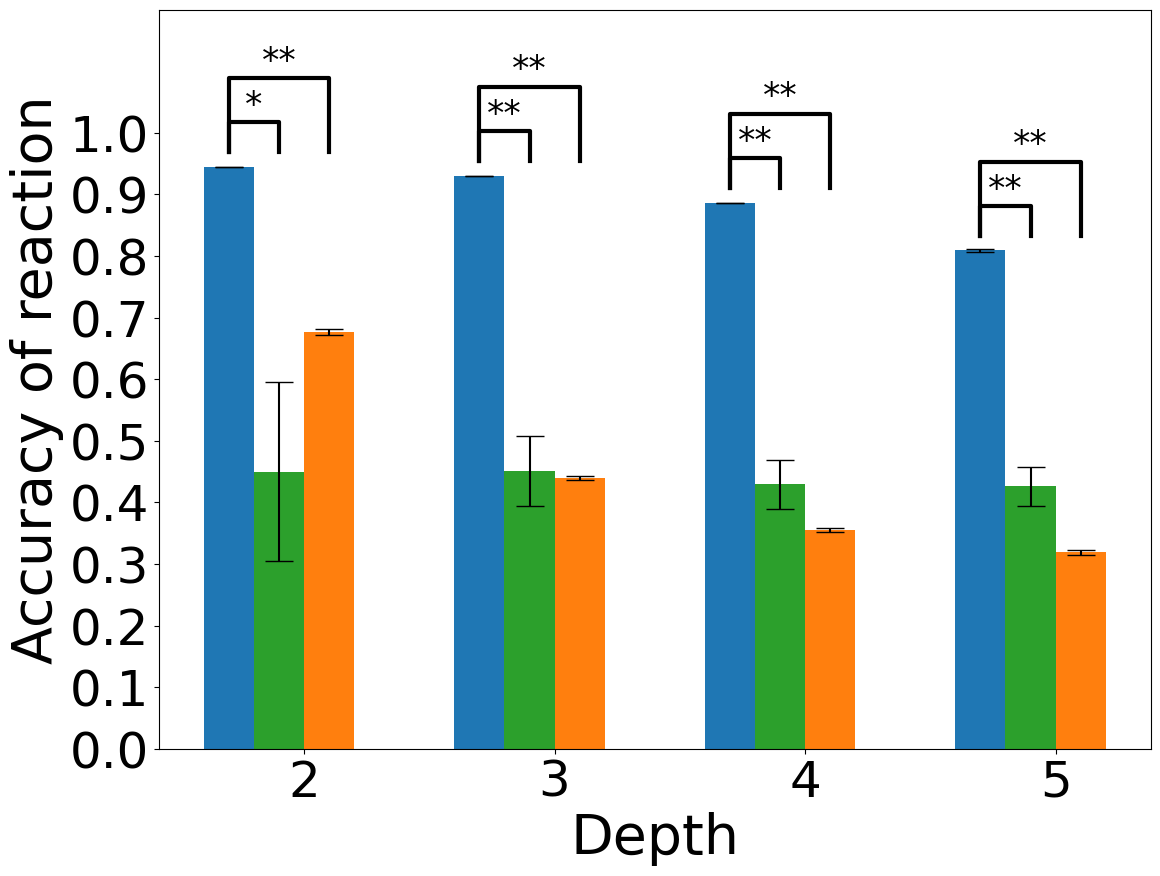}
    \vspace{-5mm}
    \subcaption{}
    \end{subfigure}
    \caption{(a) Overview of the piping inspection task environment. (b) Observation space of the robot arm. Events are registered on colored grids, and the robot arm moves between the center points of each grid (The arrow indicates the shift for framing the meter). (c) Comparisons of average returns during training in the piping inspection task with 5 random seeds. (d) Accuracy of the reaction for testing trained policies ($\ast$ : $\textit{p} < .005$, $\ast\ast$ : $\textit{p} < .0005$ by t-test). The results are averaged over 5 seeds, and the test set is the same as in Table \ref{table:test_result}.}
    \vspace{-7mm}
    \label{fig:pipe_env_result}
\end{figure}
\vspace{-1mm}
\subsection{Piping Inspection by Robot Arm}
To evaluate our algorithm in a realistic environment, we build the piping inspection task with a camera mounted on a robotic arm using Isaac Gym \cite{makoviychuk2021isaac} shown in Fig. \ref{fig:pipe_env_result} (a).
We discretized the observation space by fixing the depth motion of the arm and dividing it into a $6\times 4$ grid on a two-dimensional plane shown in Fig. \ref{fig:pipe_env_result} (b), and 10 propositions are fixed on the grid corresponding to pipe joints, a meter, and a valve to simulate the inspection.
The observation is $80\times 60$ RGB image, and action spaces of the policies and the LTL task design are the same as the navigation task.
The probability of selecting the event detector is set to 50\%, while the proposition assigned to whether the meter value is normal or abnormal is set to the uncertain disjunction.

\noindent
\textbf{Performance comparison with high-dimensional observation.} 
Fig. \ref{fig:pipe_env_result} (c) shows the performance comparisons with each method during training under the probabilistic error of the event detector.
Fig. \ref{fig:pipe_env_result} (d) presents the average reaction accuracy of behavior i) and ii) in an environment with fixed event detectors by the rollout of the learned policies.
These results show that our method has the best reactivity to uncertainty in the event detector without performance degradation, even for high-dimensional image input, suggesting the applicability of our method to real-world tasks.

As a supplement, we present an interactive inspection with a human event detector by using learned policies to show the applicability of our system for real-world applications.
We developed an operational GUI that allows a human operator to inspect the object by moving the camera with a mouse when triggered by the query policy.
See the attached video for the interactive demonstration with a human detector.

\section{DISCUSSION}
We define the event detector as a function that offers gradual evaluations of uncertainty, designed to work with human operators and edge AI systems, which use a multi-class classifier like softmax. 
Although detailed uncertainty feedback is cumbersome for humans, a workflow with several options and the corresponding uncertainty for the robot is feasible, like the demonstration video.

The limitation of this method is that the traces of tasks achieved may be biased without adjustments to the reward design since our method learns policies by sparse rewards.
A future direction to mitigate this limitation is to apply the reward shaping \cite{jothimurugan2019composable} or the multi-goal RL techniques that regularise over the goals satisfied by the agents \cite{zhao2019maximum}.

Our method leaves some topics that require further discussion as future work. First, this paper empirically shows that our method works in a setting the Possible LTL always includes the LTL progressed by true events. While this makes it possible that the policy satisfies tasks by choosing the same action as the optimal policy $\pi^*_{\Phi}$ that follows the Taskable MDP, more formal discussions on task satisfaction will be addressed in future work.
Also, our experiments do not contain the operator that can be progressed to $\mathsf{false}$, such as the until operator $\cup$, and further evaluation of the reward function is needed when the Possible LTL contains $\mathsf{false}$.
\vspace{-1mm}
\section{CONCLUSIONS}
This paper proposes a LAQBL framework for learning the agent to follow various LTL instructions with the uncertain event detectors.
The navigation task in the grid world and the inspection task with high-dimensional image input show that our method can perform in response to the uncertainty of the event detector and that effective querying of the event detector can reduce task failures due to unnecessary queries.
\addtolength{\textheight}{-12cm}   

\bibliographystyle{IEEEtran}
\bibliography{cite}

\end{document}